\documentclass[conference]{IEEEtran}
\IEEEoverridecommandlockouts
\usepackage[dvipsnames,table,xcdraw]{xcolor}

\usepackage[pagebackref=true,breaklinks=true,colorlinks,citecolor=ForestGreen]{hyperref}
\usepackage{cite}

\usepackage{amsmath,amssymb,amsfonts}
\usepackage{algorithmic}
\usepackage{graphicx}
\usepackage{textcomp}

\usepackage{booktabs}
\usepackage{url}
\usepackage{tablefootnote}
\usepackage{hyperref}
\def\BibTeX{{\rm B\kern-.05em{\sc i\kern-.025em b}\kern-.08em
    T\kern-.1667em\lower.7ex\hbox{E}\kern-.125emX}}
\begin{document}

\title{SEEC: Segmentation-Assisted Multi-Entropy Models for Learned Lossless Image Compression \thanks{$^{\star}$ Corresponding authors.}
}

\author{
\IEEEauthorblockN{Chunhang Zheng, Zichang Ren, Dou Li$^{\star}$}
\IEEEauthorblockA{School of Electronics, Peking University, Beijing 100871, China}

\IEEEauthorblockA{\{zhengch, 2101111428\}@stu.pku.edu.cn, lidou@pku.edu.cn}
}

\maketitle

\begin{abstract}
Recently, learned image compression has attracted considerable attention due to its superior performance over traditional methods. However, most existing approaches employ a single entropy model to estimate the probability distribution of pixel values across the entire image, which limits their ability to capture the diverse statistical characteristics of different semantic regions. To overcome this limitation, we propose Segmentation-Assisted Multi-Entropy Models for Lossless Image Compression (SEEC). Our framework utilizes semantic segmentation to guide the selection and adaptation of multiple entropy models, enabling more accurate probability distribution estimation for distinct semantic regions. Experimental results on benchmark datasets demonstrate that SEEC achieves state-of-the-art compression ratios while introducing only minimal encoding and decoding latency. With superior performance, the proposed model also supports Regions of Interest (ROIs) coding condition on the provided segmentation mask. Our code is available at \url{https://github.com/chunbaobao/SEEC}.
\end{abstract}

\begin{IEEEkeywords}
lossless image compression, learned image compression, entropy models, semantic segmentation
\end{IEEEkeywords}

\section{Introduction}\label{sec:introduction}

In the digital age, the demand for efficient image compression techniques has grown significantly with the exponential growth of digital images and the need for effective storage and transmission solutions. Lossless image compression, in particular, plays a crucial role in various applications, such as professional photography, medical imaging, and publications, where preserving the original image quality is paramount. Traditional image compression methods, such as PNG~\cite{boutell1997png}, JPEG-XL~\cite{alakuijala2019jpeg} and BPG~\cite{bellard2015bpg}, have made a significant impact in the past decade. However, due to the suboptimal manual design of these algorithms, the compression ratio of traditional image compression methods is often limited, leading to the need for more advanced techniques.

Recently, learning-based image compression methods have gained popularity. They have demonstrated superior performance over traditional methods in terms of compression ratio, image quality, and even speed. 
State-of-the-art learning-based lossy and lossless image compression models commonly employ context models~\cite{minnen2018joint} and hyper-prior models~\cite{balle2018variational} to capture spatial correlation information. These models extract dependencies from images and feed them into an entropy model to estimate the probability distribution of latent variables or pixel values. 
However, most existing methods are based on the assumption that the distribution can be represented by a single entropy model, which is insufficient to capture the complex and diverse characteristics of real-world images.
Generally, the prior distributions of an image vary across different semantic regions. For example, the texture of a road is typically smooth with low variance, whereas a tree exhibits complex, high-frequency patterns with significant variations in pixel intensity. Similarly, the color distribution of the sky and grass can differ significantly. In such cases, a single entropy model may not be able to accurately predict the distribution of pixel values, leading to suboptimal performance in terms of compression ratio.
\begin{figure}[t]
    \centering
    \includegraphics[width=0.95\linewidth]
    {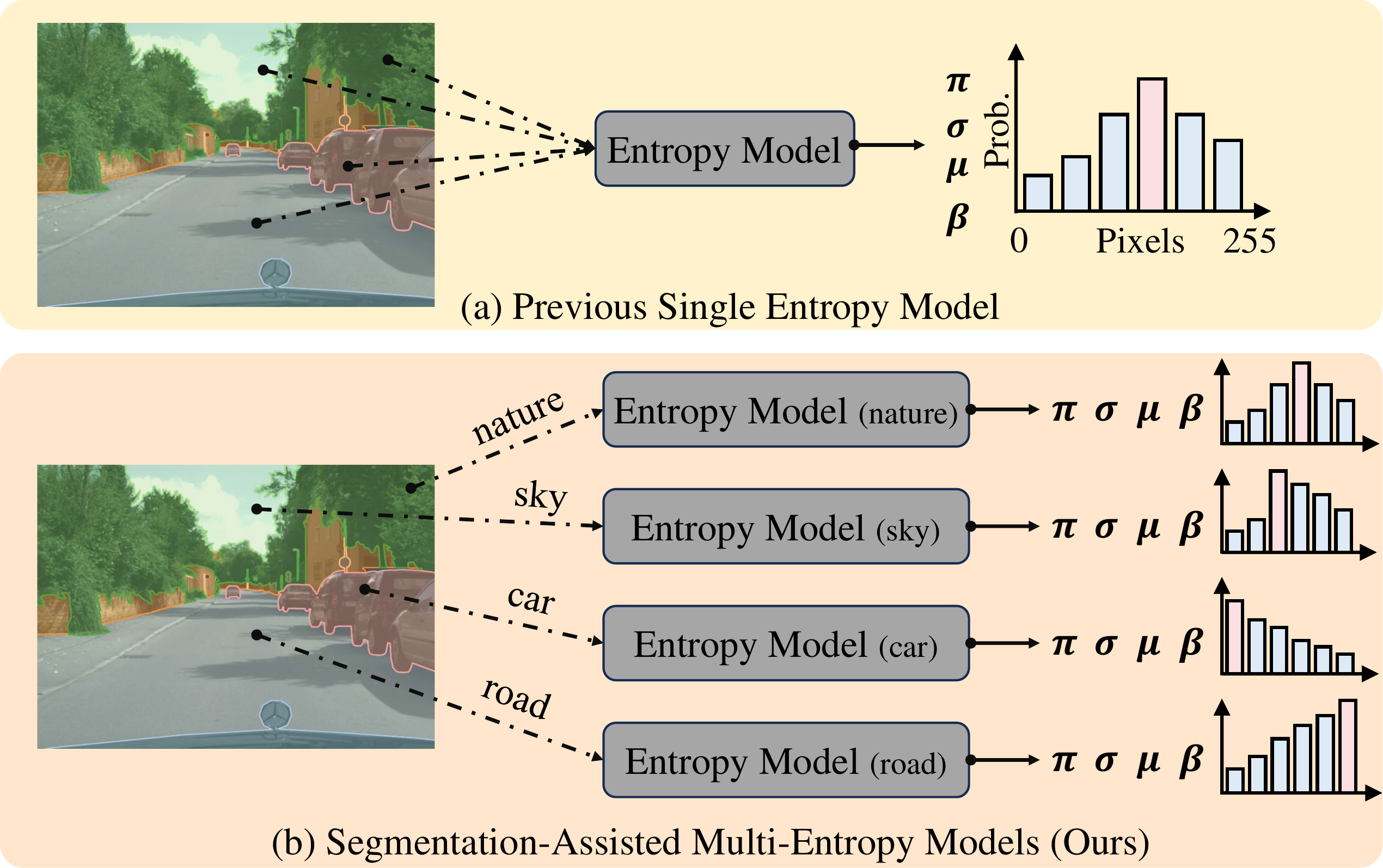}
    \caption{Entropy Model Comparison. (a) Exiting methods apply a single entropy model to the entire image. (b) Our proposed Multi-Entropy Models assign distinct entropy models to different semantic regions, allowing for more accurate modeling of pixel value distributions.}
    \label{fig:ep_comp}
\end{figure}

To address this issue, we propose a segmentation-assisted multi-entropy models for lossless image compression (SEEC) framework as shown in Figure~\ref{fig:pipeline}. In our approach, we leverage semantic segmentation to guide the selection and adaptation of multiple entropy models, allowing for more accurate predictions of pixel value distributions. Specifically, the SEEC framework first performs feature extraction from the image and encodes the latent variables into bit streams. Subsequently, semantic segmentation is applied to identify different semantic regions within the image. Each region is then assigned a specific entropy model based on its unique characteristics, enabling the framework to capture the parameters of diverse distributions more effectively. Finally, the probability distributions of the pixel values are modeled using a multi-channel discrete logistic mixture likelihood, where the parameters of the mixture components are predicted by the corresponding entropy model. Our approach has been evaluated on several benchmark datasets.
% , including DIV2K~\cite{agustsson2017ntire}, Kodak~\cite{kodak}, Adobe's Portrait Segmentation~\cite{shen2016automatic},  and CLIC~\cite{clic2020dataset} 
The results show that our SEEC framework, with segmentation assistance, achieves state-of-the-art compression performance. Our contributions are summarized as follows:
\begin{itemize}
    \item We propose a segmentation-assisted multi-entropy model for lossless image compression, leveraging semantic segmentation to guide the selection and adaptation of multiple entropy models. This approach enables more precise predictions of pixel value distributions, significantly enhancing compression performance.
    \item We design a multi-channel discrete logistic mixture likelihood to predict the mixture coefficients for each channel of the image. This design allows our framework to effectively capture the parameters of diverse distributions, improving modeling accuracy.
    % \item Extensive experiments on multiple benchmark datasets demonstrate that our SEEC framework achieves state-of-the-art compression performance. 
    \item  We propose a regions of interest (ROIs) coding strategy based on the segmentation masks, allowing for more efficient compression while maintaining lossless reconstruction of the interesting regions.

\end{itemize}
\section{Related Works}\label{sec:related-works}

\subsection{Learned Lossy Image Compression}

Similar to traditional lossy image codecs, learned lossy image compression methods also rely on three main steps: transform, quantization, and entropy coding. However, instead of using hand-crafted transforms, learned lossy image compression methods utilize convolutional neural networks to learn an optimal nonlinear transform that maps input images into a compact latent representation~\cite{balle2016end, balle2018variational, minnen2018joint}, which can be considered as a case of Variational Auto-Encoder (VAE)~\cite{kingma2013auto}. 
This latent representation is then quantized, often using techniques like uniform quantization~\cite{balle2016end} to deal with the non-differentiability of quantization during training. After quantization, entropy coding are applied to further compress the representation into a bit-stream. 

Learned lossy image compression methods are typically trained end-to-end using a combination of reconstruction loss and rate loss to optimize the trade-off between compression performance and image quality:
\begin{equation}
    \mathcal{L} = \mathcal{R} + \lambda \mathcal{D},
\end{equation}
where $\mathcal{R}$ is the bit rate of the latent representation, $\mathcal{D}$ is the distortion between the original and reconstructed image, and $\lambda$ is a hyperparameter that controls the trade-off between compression rate and image quality.

% The hyper-prior model~\cite{balle2018variational} and the context model~\cite{minnen2018joint} are widely adopted techniques in learned lossy image compression for improving entropy estimation. The hyper-prior model extracts global spatial information by analyzing a secondary set of latent variables (hyper-latents), while the context model captures local dependencies through autoregressive context from neighboring latent elements. The outputs of both the hyper-prior and context model are fed into a single entropy model to estimate the parameters of the probability distribution of the latent variables. These approaches allow for more accurate modeling of the latent variable distributions, leading to improved compression performance.

\subsection{Learned Lossless Image Compression}
Different from learned lossy image codecs, where the distortion between the original image and the reconstructed image needed to be optimized, learned lossless image compression models directly estimate the probability distribution of the pixel values. The main goal of these models is to minimize the bit rate of the latent variables and pixel values:
\begin{equation}\label{eq:tworate}
    \mathcal{L}=\mathcal{R}=\mathcal{R}_\text{latent}+\mathcal{R}_\text{pixel},
\end{equation}

The compression performance of learned lossless image compression models fundamentally relies on the accuracy of the estimated probability distribution of pixel values. To achieve this, various neural network architectures have been proposed to model the dependencies among pixels effectively. These models are typically classified into three categories: autoregressive models~\cite{salimans2017pixelcnn++}, VAE models~\cite{ mentzer2019practical,bai2024deep}, and flow-based models~\cite{zhang2021ivpf,zhang2021iflow}.

\section{Methodology}\label{sec:methodologies}

% \subsection{Motivation}

% Different semantic regions within an image exhibit distinct statistical characteristics due to variations in texture, color, and structure. We illustrate the distribution of pixel values across several semantic regions from the Cityscapes~\cite{Cordts2016Cityscapes} dataset in Figure~\ref{fig:rgb_comp}. It is evident that the pixel values of different semantic regions are not identically distributed, indicating that a single entropy model may struggle to capture such heterogeneity effectively. Therefore, we argue that using multiple entropy models is necessary.

\begin{figure*}[!t] 
    \centering 
    \includegraphics[width=0.95\linewidth] 
    {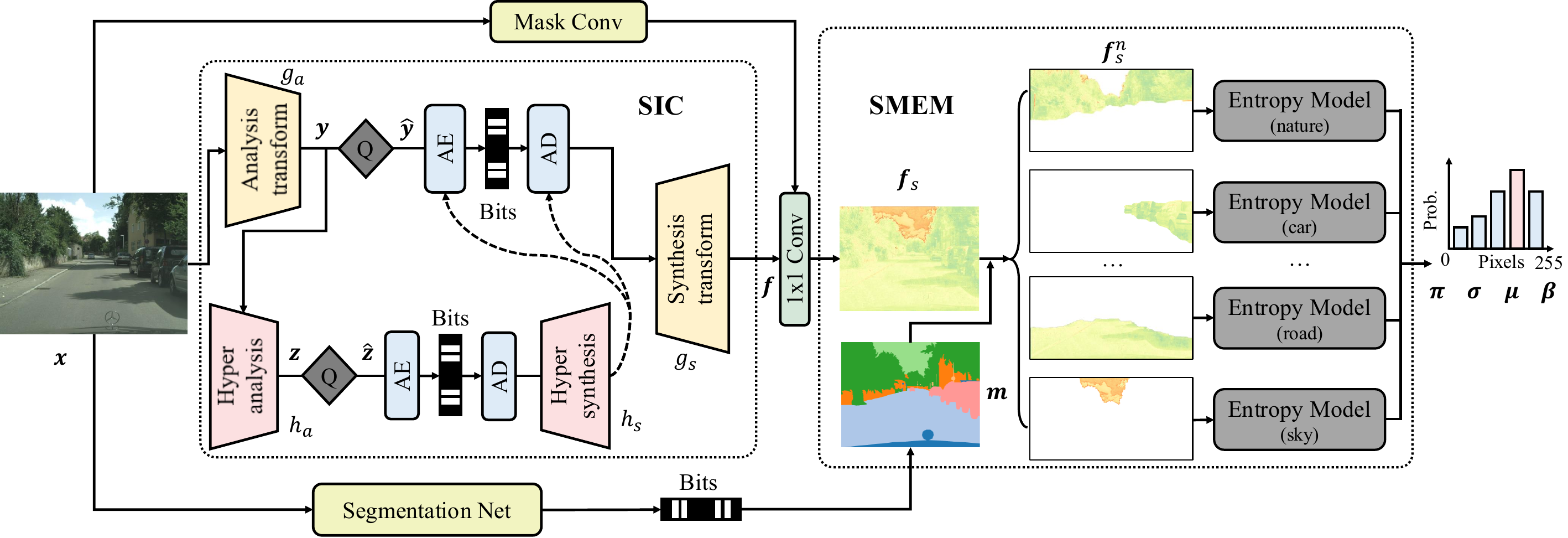}
    \caption{The overall architecture of SEEC. The left part shows the Segmentation-aware Image Compressor (SIC) module, which extracts segmentation-aware features from the input image. The right part illustrates the Segmentation-assisted Multi-Entropy Models (SMEM) module, which estimates the probability distribution of pixel values using multiple entropy models guided by semantic segmentation. Q denotes the quantization operation. AE and AD represent arithmetic encoder and arithmetic decoder, respectively.}
    \label{fig:pipeline}
\end{figure*} 

\subsection{Segmentation-assisted Multi-Entropy Models}

To capture the heterogeneous statistics across different semantic regions, we propose segmentation-assisted multi-entropy model framework for lossless image compression (SEEC), as shown in Figure~\ref{fig:pipeline}. SEEC consists of two main components: Segmentation-aware Image Compressor (SIC) and Segmentation-assisted Multi-Entropy Models (SMEM).
The SIC module compresses the input image into a latent representation while extracting segmentation-aware features that facilitate the subsequent entropy modeling process. 
Meanwhile, a segmentation net is employed to generate semantic segmentation masks.
Then, the SMEM module estimates the probability distribution of pixel values through multiple entropy models, guided by semantic segmentation.

In SIC, we adopt ELIC framework~\cite{he2022elic} as the backbone architecture.
Specifically, the input image $\boldsymbol{x} \in \mathbb{R}^{3 \times H \times W}$ is first transformed into a latent representation using an analysis transform $\boldsymbol{y} = g_a(\boldsymbol{x})$. This quantized latent is then passed through a hyper analysis to generate side information $\boldsymbol{z} = h_a(\boldsymbol{\hat{y}})$, which is used to estimate the distribution parameters of the quantized latent representation $\boldsymbol{\hat{y}}$. The quantized latent representation $\boldsymbol{\hat{y}}$ is assumed to follow a Gaussian distribution with mean and variance predicted by the hyper synthesis $\boldsymbol{\mu}, \boldsymbol{\sigma} = h_s(\boldsymbol{\hat{z}})$. Reversely, features are extracted from the quantized latent representation using a synthesis transform $\boldsymbol{f} = g_s(\boldsymbol{\hat{y}})$. During training, quantization is approximated by adding uniform noise to enable end-to-end optimization via gradient descent.

% \begin{equation}
%     \mathcal{R}_{\text{latent}} = \mathcal{R}_y + \mathcal{R}_z,
% \end{equation}
% \begin{equation}
%     \mathcal{R}_y = \mathbb{E}_{p(\boldsymbol{x})} \, \mathbb{E}_{q_{\theta}(\tilde{\boldsymbol{y}} \mid \boldsymbol{x})} \, \mathbb{E}_{q_{\phi}(\tilde{\boldsymbol{z}} \mid \tilde{\boldsymbol{y}}, \boldsymbol{x})} \left[ - \log_2 p_{\varphi}(\tilde{\boldsymbol{y}} \mid \tilde{\boldsymbol{z}}) \right], \label{eq:yrate}
% \end{equation}
% \begin{equation}
%     \mathcal{R}_z = \mathbb{E}_{p(\boldsymbol{x})} \, \mathbb{E}_{q_{\theta}(\tilde{\boldsymbol{y}} \mid \boldsymbol{x})} \, \mathbb{E}_{q_{\phi}(\tilde{\boldsymbol{z}} \mid \tilde{\boldsymbol{y}}, \boldsymbol{x})} \left[ - \log_2 p_{\phi}(\tilde{\boldsymbol{z}} \mid \tilde{\boldsymbol{y}}) \right], \label{eq:zrate}
% \end{equation}
% where $\theta$, $\phi$ and $\varphi$ are the parameters of the analysis transform, synthesis transform and hyper synthesis, respectively.

In SMEM, given the features $\boldsymbol{f}$ generated from the latent representation, we apply an autoregressive model~\cite{minnen2018joint} to extract the pixel-wise context features $\boldsymbol{C}_{\boldsymbol{x}}$ from the original input $\boldsymbol{x}$. Both the latent features $\boldsymbol{f}$ and the extracted context features $\boldsymbol{C}_{\boldsymbol{x}}$ are then fused to generate the segmentation-aware features $\boldsymbol{f}_s$, which are subsequently used for segmentation-assisted entropy modeling. Therefore, the distribution of the input image $\boldsymbol{x}$ is modeled as
\begin{equation}
   p(\boldsymbol{x} \mid \boldsymbol{C}_{\boldsymbol{x}}, \boldsymbol{f})= 
    \prod_{i=1}^{H \times W} p(x_i \mid x_{<i}, \boldsymbol{f}) = p_{\theta_{\text{me}}} (\boldsymbol{x} \mid \boldsymbol{f}_s) , \label{ep:thetame}
\end{equation}

where $x_{<i}$ denotes the pixels preceding $x_i$ in the raster scan order, typically modeled using masked convolutional layers~\cite{van2016conditional, salimans2017pixelcnn++}. The $\theta_\text{me}$ denotes the parameters of the multiple entropy models. 

To effectively capture the heterogeneous statistics of various semantic regions within an image, we assign each region to a dedicated entropy model. Specifically, the segmentation-aware features $\boldsymbol{f}_s$ are first partitioned into $N$ non-overlapping groups based on semantic segmentation masks, where each group corresponds to a distinct semantic class. For the $n$-th semantic class, the associated features $\boldsymbol{f}_s^n$ are computed as:

\begin{equation} 
    \boldsymbol{f}_s^n = \boldsymbol{f}_s \odot \boldsymbol{m}^n, \quad n = 0, 1, \dots, N - 1 , \label{eq:segmentation}
\end{equation}
where $\boldsymbol{m}^n$ denotes the binary segmentation mask for the $n$-th class, and $\odot$ represents element-wise multiplication. These class-specific features $\boldsymbol{f}_s^n$ are then individually processed by multiple entropy models. The multiple entropy models can be represented as a set of single entropy models $\theta_{\text{me}}^n$, each specialized for modeling the distribution of pixel values in a specific semantic region. The overall probability distribution of the input image $\boldsymbol{x}$ is thus expressed as a mixture of $N$ single entropy models, where the contribution of each model is modulated by its associated segmentation mask. Combining Equations~\eqref{ep:thetame} and~\eqref{eq:segmentation}, the full distribution is given by:
\begin{equation}
    p_{\theta_{\text{me}}}(\boldsymbol{x} \mid \boldsymbol{f}_s) = \sum_{n=0}^{N-1} \boldsymbol{m}^n \cdot p_{\theta_{\text{me}}^n}(\boldsymbol{x} \mid \boldsymbol{f}_s^n), \label{ep:pme}
\end{equation}

\subsection{Multi-channel Discrete Logistic Mixture Likelihood}

To model the complex distribution of pixel values in natural images, based on the discrete logistic mixture likelihood~\cite{salimans2017pixelcnn++}, we propose a multi-channel discrete logistic mixture likelihood. For an RGB image $\boldsymbol{x} \in \{0, 1, \ldots, 2^{l} - 1\}^{H \times W}$ with $l$-bit pixel values, we have
\begin{equation}
    p_{\theta_{me}^n}(\boldsymbol{x} \mid \boldsymbol{f}_s^n) =\prod_{i=1}^{H \times W} p_{\theta_{me}^n}(x_\text{r}^i, x_\text{g}^i, x_\text{b}^i \mid \boldsymbol{f}_s^{ni}) 
\end{equation} 
where $x_\text{r}^i$, $x_\text{g}^i$, $x_\text{b}^i$ are the pixel values of the $i$-th pixel in the red, green, and blue channels, respectively. 
And $\boldsymbol{f}_s^{ni}$ is the segmentation-aware feature for the $i$-th pixel in the $n$-th semantic region. Channel autoregressive modeling ~\cite{salimans2017pixelcnn++} is applied to model the distribution of pixel values in each channel:
\begin{align}
    &p_{\theta_{me}^n}(x_\text{r}^i, x_\text{g}^i, x_\text{b}^i \mid \boldsymbol{f}_s^{ni}) 
    = \nonumber \\
    &p_{\theta_{me}^n}(x_\text{r}^i \mid \boldsymbol{f}_s^{ni}) 
    \cdot p_{\theta_{me}^n}(x_\text{g}^i\mid x_\text{r}^i, \boldsymbol{f}_s^{ni}) 
    \cdot p_{\theta_{me}^n}(x_\text{b}^i \mid x_\text{r}^i, x_\text{g}^i, \boldsymbol{f}_s^{ni}).
\end{align}

Similar to the discrete logistic mixture likelihood~\cite{salimans2017pixelcnn++}, our model defines the distribution parameters as the means $\boldsymbol{\mu}_c^{ik}$, variances $\boldsymbol{s}_c^{ik}$, mixture coefficients $\boldsymbol{\beta}_c^{ik}$, and mixture weights $\boldsymbol{\pi}_c^{ik}$ for each channel $c \in \{r, g, b\}$ and mixture component $k \in \{1, 2, \dots, K\}$. Unlike the original discrete logistic mixture likelihood, which uses shared mixture weights $\boldsymbol{\pi}^{ik}$ across all channels, we assign channel-specific mixture weights $\boldsymbol{\pi}_c^{ik}$ to better capture the distinct statistical properties of different semantic regions in each channel.  
The autoregressive modeling is reflected in the means for each channel:
\begin{align}
    \boldsymbol{\hat{\mu}}_r^{ik} &= \boldsymbol{\mu}_r^{ik}, \quad
    \boldsymbol{\hat{\mu}}_g^{ik} = \boldsymbol{\mu}_g^{ik} + \boldsymbol{\beta}_r^{ik} \cdot x_r^{i},  \nonumber \\
    \boldsymbol{\hat{\mu}}_b^{ik} &= \boldsymbol{\mu}_b^{ik} + \boldsymbol{\beta}_g^{ik} \cdot x_r^{i} + \boldsymbol{\beta}_b^{ik} \cdot x_g^{i}.
\end{align}

\subsection{ROIs Coding}

Most learned lossless image compression methods treat all pixels equally, ignoring perceptual importance. To address this, we extend SEEC to support ROI-based coding. Foreground regions (ROIs) are preserved losslessly, while background regions (non-ROIs) are reconstructed with relaxed fidelity to save bits. Formally,
\begin{align}
    \boldsymbol{x}_\text{r} &= \boldsymbol{x} \odot \boldsymbol{m}^1, &
    \boldsymbol{x}_\text{n} &= \boldsymbol{x} \odot \boldsymbol{m}^0,
\end{align}
where $\boldsymbol{m}^1$ and $\boldsymbol{m}^0$ are the binary masks for ROIs and non-ROIs. ROIs $\boldsymbol{x}_\text{r}$ follow standard lossless compression, while non-ROIs $\boldsymbol{x}_\text{n}$ skip entropy encoding and are reconstructed via maximum likelihood sampling:
\begin{equation}
\boldsymbol{\hat{x}}_\text{n} = \arg\max{\boldsymbol{x}} {p_{\theta_{\text{me}}^0}(\boldsymbol{x} \mid \boldsymbol{f}_s^0)},
\end{equation}
where $p_{\theta_{\text{me}}^0}$ is the background entropy model and $\boldsymbol{f}_s^0$ are segmentation-aware background features.
\section{Experiments}\label{sec:experiments}

\subsection{Settings}\label{subsec:settings}

\begin{table*}[!t]
    \centering
    \caption{Lossless image compression performance (bpp) of our proposed method compared to other lossless image codec on DIV2K, Adobe Portrait, Urban100, CLIC.mobile, Kodak, and Cityscapes datasets. The top results are highlighted in \textbf{bold}.}
    \label{tab:bpp_results}
    \renewcommand{\arraystretch}{1.15}
    \setlength{\tabcolsep}{12pt}
    \begin{tabular*}{0.95\textwidth}{@{\extracolsep{\fill}}cccccccc@{}}
      \toprule
        Codec & DIV2K & Adobe Portrait & Urban100 & CLIC.m & Kodak & Cityscapes & ~ \\ 
        \midrule
        QOI~\cite{qoi} & 14.34 & 11.32 & 16.46 & 13.10 & 13.99 & 10.60 & ~ \\ 
        BPG~\cite{bellard2015bpg} & 13.26 & 11.30 & 14.86 & 12.71 & 14.08 & 8.95 & ~ \\ 
        PNG~\cite{boutell1997png} & 12.82 & 8.74 & 14.33 & 11.64 & 13.44 & 8.94 & ~ \\ 
        JPEG-LS~\cite{weinberger2000loco} & 11.96 & 9.51 & 13.72 & 11.54 & 13.07 & 7.30 & ~ \\ 
        JPEG2000~\cite{skodras2001jpeg} & 9.38 & 5.72 & 10.80 & 8.16 & 9.58 & 6.83 & ~ \\ 
        WebP~\cite{webp} & 9.53 & 5.45 & 10.77 & 8.32 & 9.62 & 6.67 & ~ \\ 
        JPEG-XL~\cite{alakuijala2019jpeg} & 8.68 & 4.92 & 9.71 & 7.45 & 9.18 & 6.08 & ~ \\ 
        FLIF~\cite{sneyers2016flif} & 8.73 & 4.86 & 9.80 & 7.44 & 9.04 & 6.11 & ~ \\ 
        \midrule
        L3C~\cite{mentzer2019practical} & 9.27 & 6.20 & 10.96 & 7.92 & 9.78 & - & ~ \\ 
        RC~\cite{mentzer2020learning} & 9.24 & - & - & 7.62 & - & - & ~ \\ 
        SReC~\cite{cao2020lossless} & 8.47 & 5.77 & 9.92 & 7.32 & 9.10 & 6.05 & ~ \\ 
        Near-Lossless~\cite{bai2021learning} & 8.43 & - & - & 7.53 & 9.12 & - & ~ \\ 
        iVPF~\cite{zhang2021ivpf} & 8.04 & - & - & 7.17 & - & - & ~ \\ 
        iFlow~\cite{zhang2021iflow} & 7.71 & - & - & 6.78 & - & - & ~ \\ 
        MS-CFA~\cite{li2025deep} & 7.62 & - & - & 6.98 & 8.97 & - & ~ \\ 
        ArIB-BPS~\cite{zhang2024learned} & 7.65 & - & - & - & - & - & ~ \\ 
        DLPR~\cite{bai2024deep} & 7.65 & 4.45 & 9.09 & 6.48 & 8.58 & 5.51 & ~ \\ 
        \midrule
        SEEC (Ours)\tablefootnote{Note that our method includes the bitrates for storing segmentation masks.}
 & \textbf{7.54} & \textbf{4.28} & \textbf{8.87} & \textbf{6.41} & \textbf{8.53} & \textbf{5.43} &  \\ 
      \bottomrule
    \end{tabular*}
  \end{table*}
We train our SEEC model on the DIV2K dataset~\cite{ignatov2018pirm} consisting 800 high-quality images. During training, each image is fed into segmentation model to generate the corresponding segmentation masks. Both the images and their corresponding segmentation masks are initially split into non-overlapping $128 \times 128$ patches. We apply horizontal and vertical flipping with a probability of $0.5$, then randomly crop the patches to $64 \times 64$. The model is trained for $1500$ epochs using the Adam optimizer~\cite{kingma2014adam} with a batch size of $64$. The learning rate is initialized at $1 \times 10^{-4}$ and adaptively reduced by a factor of $0.9$ if the validation loss does not improve for $30$ consecutive epoch. We set the number of classes to $N=2$ as described in section~\ref{subsubsec:seg_analysis}, and employ BiRefNet~\cite{zheng2024bilateral} as the segmentation model. The segmentation masks are stored using JPEG-XL.

We conduct a comprehensive evaluation across six benchmark datasets representing diverse image characteristics:
\begin{itemize}
    \item \textbf{DIV2K}~\cite{ignatov2018pirm}: Validation set of DIV2K dataset, containing 100 2K resolution images.
        \item \textbf{Adobe's Portrait}~\cite{shen2016automatic}: A dataset with 1800 portrait images and corresponding segmentation masks, which is originally used for semantic segmentation tasks.
    \item \textbf{Urban100}~\cite{Huang_2015_CVPR}: A dataset with 100 high-resolution urban images, which is originally used for image super-resolution tasks.
    \item \textbf{CLIC.mobile}~\cite{clic2020dataset}: CLIC mobile validation dataset with 61 2K resolution images.
    \item \textbf{Kodak}~\cite{kodak}: A widely used benchmark dataset containing 24 768$\times$512 color images.

    \item \textbf{Cityscapes}~\cite{Cordts2016Cityscapes}: A dataset with 5000 high-resolution images of urban street scenes.
\end{itemize}

The implementation of the model is based on CompressAI~\cite{begaint2020compressai}. We train the SEEC model on NVIDIA A100 GPU, while evaluate the compression performance and running time on Intel CPU i7-12700K, 32 G RAM and NVIDIA RTX3090 GPU.

\subsection{Coding Performance of SEEC}

To evaluate the effectiveness of our proposed SEEC codec, we compare it against eight traditional lossless image compression methods, including QOI~\cite{qoi}, BPG~\cite{bellard2015bpg}, PNG~\cite{boutell1997png}, JPEG-LS~\cite{weinberger2000loco}, JPEG2000~\cite{skodras2001jpeg}, WebP~\cite{webp}, JPEG-XL~\cite{alakuijala2019jpeg}, and FLIF~\cite{sneyers2016flif},
and nine recent learned lossless image compression methods, including L3C~\cite{mentzer2019practical}, 
RC~\cite{mentzer2020learning}, SReC~\cite{cao2020lossless}, 
% Bit-Swap~\cite{kingma2019bit}, 
% HiLLoC~\cite{townsend2019hilloc}, 
% IDF~\cite{hoogeboom2019integer}, IDF++~\cite{berg2020idf++}, 
% LBB~\cite{ho2019compression}, 
iVPF~\cite{zhang2021ivpf}, iFlow~\cite{zhang2021iflow}, MS-CFA~\cite{li2025deep},
Near-Lossless~\cite{bai2021learning}, ArIB-BPS~\cite{zhang2024learned}, and DLPR~\cite{bai2024deep}.
For fair comparison, we exclude LLM-based methods and overfitting-based methods due to their excessive encoding or decoding time. 
All methods are evaluated in terms of bits per pixel (bpp) on the corresponding test sets, as shown in Table~\ref{tab:bpp_results}.

Even with the additional overhead of storing segmentation masks, SEEC consistently outperforms both traditional and learned compression baselines.
Compared with the best traditional method, FLIF, SEEC achieves a bitrate reduction at average of 11.0\% across all datasets.
Our approach achieves 4.28 bpp on Adobe's Portrait dataset, outperforming the previous best DLPR (4.45 bpp) by 3.9\%

Notably, SEEC adopts the same analysis transform, synthesis transform, hyper analysis, and hyper synthesis modules as DLPR, suggesting that the performance gain primarily stems from our Segmentation-assisted Multi-Entropy Models, underscoring the effectiveness of semantic-aware compression.

\subsection{Efficiency of Codec}

To evaluate the efficiency of our SEEC codec, we measure the average runtime (in seconds) required to encode and decode a single image. We compare the runtime of our SEEC framework with several representative lossless image codecs, including WebP~\cite{webp}, JPEG-XL~\cite{alakuijala2019jpeg}, BPG~\cite{bellard2015bpg}, FLIF~\cite{sneyers2016flif},
L3C~\cite{mentzer2019practical}, SReC~\cite{cao2020lossless}, Minnen~\cite{minnen2018joint}, ArIB-BPS~\cite{zhang2024learned}, and DLPR~\cite{bai2024deep}. We also include a variant of SEEC with only single entropy model (denoted as SEEC sig.) to demonstrate the efficiency of our multi-entropy models.
The results are summarized in Table~\ref{tab:efficiency}, show that the multi-entropy models incur only marginal additional encoding and decoding time.

\begin{table}[htbp]
  \centering
  \renewcommand{\arraystretch}{1.1}
  \setlength{\tabcolsep}{8pt}
  \caption{Average runtime (sec.) of SEEC compared to other codecs (Encoding/Decoding time).}

  \begin{tabular}{lcccc}
  \toprule
  Codec & 768$\times$512 & 1024$\times$768  &  2048$\times$1536\\
  \midrule
WebP~\cite{webp} & 0.15/0.01 & 0.2/0.01 & 2.23/0.05 \\
JPEG-XL~\cite{alakuijala2019jpeg} & 0.24/0.06 & 0.39/0.09 & 2.97/0.53 \\
BPG~\cite{bellard2015bpg} & 0.23/0.19 & 0.44/0.32 & 5.71/1.42 \\
FLIF~\cite{sneyers2016flif} & 4.13/0.66 & 7.31/1.13 & 33.07/5.27 \\

SReC~\cite{cao2020lossless} & 0.58/0.59 & 1.10/1.16 & 4.11/4.58 \\

L3C~\cite{mentzer2019practical} & 0.70/0.64 & 1.30/1.26 & 4.93/5.06 \\
Minnen~\cite{minnen2018joint} & 1.26/2.50 & 2.48/5.01  & 9.97/19.81  \\
ArIB-BPS~\cite{zhang2024learned} & 6.59/6.42 & OOM & OOM\tablefootnote{OOM: out of memory}
 \\
  DLPR ~\cite{bai2024deep} & 0.97/1.49 & 1.68 /2.76 & 6.96 /11.27 \\
  \midrule
SEEC (sig.) &1.10/1.45 & 1.88 /2.73 & 7.07/10.74 \\
SEEC (Ours)\tablefootnote{Note that our method includes the overhead of the inference time of the segmentation model and mask encoding/decoding.} & 1.11/1.46 & 1.91 /2.74 & 7.31/10.97 \\

  \bottomrule

\end{tabular}
\label{tab:efficiency}
\end{table}

\subsection{ROIs Coding}

\begin{figure}
    \centering
    \includegraphics[width=1\linewidth]
    {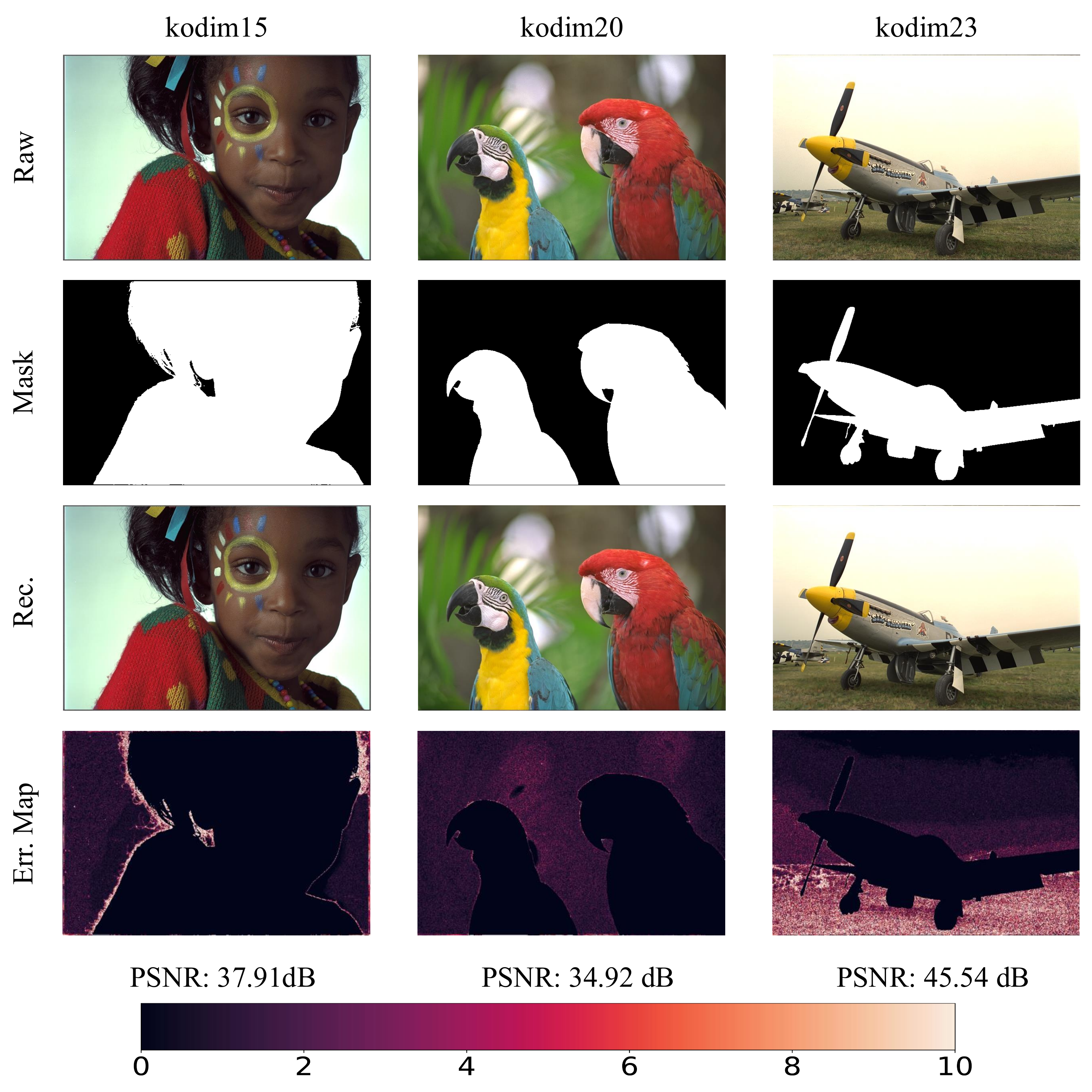}
    \caption{Visualization of ROIs coding. From top to bottom: original image, segmentation mask where white regions denote the regions of interest (foreground), reconstructed image using ROIs coding, and error map where brighter pixels indicate larger reconstruction errors.}
    \label{fig:roi}

\end{figure}

See Figure~\ref{fig:roi} for visual examples of ROIs coding. By skipping the entropy coding stage for non-ROIs, SEEC achieves lower bit rates while preserving lossless reconstruction within the ROIs. Compared to full-image coding, ROIs coding reduces runtime by 25\%, since entropy coding and decoding are applied only to the ROIs, and cuts bit rates by 50\% by omitting non-ROIs from encoding.

\subsection{Ablation Studies}

\subsubsection{Analysis of Network Components}

To further evaluate the contributions of the segmentation-assisted multi-entropy models and the multi-channel discrete logistic mixture distributions, we conduct ablation studies on SEEC with different components.
We train our models with hyper-prior architecture for 600 epochs, with the learning rate initialized to $1 \times 10^{-4}$ and decayed by a factor of 0.9 at epochs $[350, 390, 430, 470, 510, 550, 590]$. The results, summarized in Table~\ref{tab:ablation}, show that both the segmentation-assisted multi-entropy models and the multi-channel discrete logistic mixture distributions substantially contribute to the performance improvements of SEEC.

\begin{table}[h]
  \centering
  \renewcommand{\arraystretch}{1.1}
  \setlength{\tabcolsep}{5pt}
  \small 
  \caption{Ablation study of SEEC on  CLIC.m, Urban100 datasets and DIV2K. SMEM denotes the segmentation-assisted multi-entropy models, and MCDLM denotes the multi-channel discrete logistic mixture distributions.}
  \begin{tabular}{cc|ccc}
  \toprule
  SMEM & MCDLM  & CLIC.m & Urban100 & DIV2K\\
  \midrule
  $\times$ & $\times$  & 6.64 (+0.15) & 9.22 (+0.19) & 7.73 (+0.11) \\
  $\checkmark$ & $\times$  & 6.55 (+0.06) & 9.11 (+0.08) & 7.68 (+0.06) \\
  $\times$ & $\checkmark$  & 6.59 (+0.10) & 9.15 (+0.12) & 7.70 (+0.08) \\
  \midrule
  $\checkmark$ & $\checkmark$  & 6.49 & 9.03 & 7.62 \\
  \bottomrule
\end{tabular}

\label{tab:ablation}

\end{table}

\subsubsection{Analysis of the number of classes} 
To examine how the number of segmentation classes affects compression performance, we train SEEC models with different class counts $N = 2$, $N = 4$, $N = 6$, and $N = 8$ on the Cityscapes dataset using teacher-forced segmentation masks. The results are presented in Table~\ref{tab:num_cls}. As $N$ increases, the bit rate for image content $BPP_{x}$ decreases, indicating improved compression efficiency, while the bit rate for storing segmentation masks $BPP_{seg}$ increases. Despite this, the total bit rate $BPP_{total}$ slightly decreases overall. Higher values of $N$ also lead to modestly longer encoding and decoding times due to the increased complexity of handling multiple entropy models, representing the primary trade-off. Considering these factors, and the fact that each image contains at least two main semantic regions (foreground and background), we choose $N = 2$ in our main experiments to balance compression performance and computational efficiency.

\begin{table}[h]
  \centering
  \renewcommand{\arraystretch}{1.1}
  \setlength{\tabcolsep}{5pt}
  \small 
  \caption{Impact of the number of segmentation classes on performance on Cityscapes dataset.}
  \begin{tabular}{c|cccc}
  \toprule
  N  & $BPP_{total}$ & $BPP_{x}$ & $BPP_{seg}$ & $Time_{enc/dec}$ \\
  \midrule
  2  & 6.18 & 6.09 & 0.0054 & 4.94/7.31 \\
  4  & 6.15 & 6.07 & 0.0170 & 5.32/7.68 \\
  6  & 6.13 & 5.98 & 0.0194 & 5.70/8.05 \\
  8  & 6.11 & 5.96 & 0.0199 & 5.96/8.34 \\

  \bottomrule
\end{tabular}

\label{tab:num_cls}
\end{table}

\subsubsection{Analysis of Segmentation}\label{subsubsec:seg_analysis}

To evaluate the overhead from the segmentation model and mask storage, we test on the Kodak dataset. The segmentation model takes 0.14 seconds per image, while JPEG-XL encoding/decoding of masks takes 0.06 seconds and 0.01 seconds, respectively. The masks add only 0.02 bpp on average, 
marginal compared to the total 8.53 bpp. These results confirm that SEEC introduces minimal runtime and bit-rate overhead, remaining practical for semantic-aware lossless compression.
\section{Conclusion}\label{sec:conclusion}
In this paper, we introduce SEEC, a framework that leverages semantic segmentation to model the heterogeneous statistics of different regions. By assigning specialized entropy models, SEEC improves probability estimation and compression efficiency. Experiments demonstrate state-of-the-art compression with minimal latency. Moreover, SEEC naturally extends to ROI-based coding, using segmentation masks for selective compression. These results highlight the effectiveness of integrating semantic information into learned lossless image compression.
\section{ACKNOWLEDGEMENTS}

This work is supported by National Key R\&D Program of China under Award Numbers 2020YFB1807802, 2016ZX03001018-005.

\bibliographystyle{IEEEbib}
\bibliography{icme2026references}

\end{document}